# Classification of Two-channel Signals by Means of Genetic Programming

Daniel Rivero
University of A Coruna
Facultade de Informatica
Campus de Elviña 15071
A Coruña Spain
0034 881016040
drivero@udc.es

Enrique Fernandez-Blanco
University of A Coruna
Facultade de Informatica
Campus de Elviña 15071
A Coruña Spain
0034 881011302
efernandez@udc.es

Julian Dorado
University of A Coruna
Facultade de Informatica
Campus de Elviña 15071
A Coruña Spain
0034 881011239
julian@udc.es

Alejandro Pazos
University of A Coruna
Facultade de Informatica
Campus de Elviña 15071
A Coruña Spain
0034 881011231
apazos@udc.es

## ABSTRACT

*Traditionally, signal classification is a process in which previous knowledge of the signals is needed. Human experts decide which features are extracted from the signals, and used as inputs to the classification system. This requirement can make significant unknown information of the signal be missed by the experts and not be included in the features. This paper proposes a new method that automatically analyses the signals and extracts the features without any human participation. Therefore, there is no need of previous knowledge about the signals to be classified. The proposed method is based on Genetic Programming and, in order to test this method, it has been applied to a well-known EEG database related to epilepsy, a disease suffered by millions of people. As the results section shows, high accuracies in classification are obtained.*

## Categories and Subject Descriptors

I.5.2 [Design Methodology]: *Classifier design and evaluation, Feature evaluation and selection, Pattern analysis*

I.5.4 [Applications]: *Signal processing*

## General Terms

Algorithms, Measurement, Performance, Experimentation.

## Keywords

Genetic Programming, Evolutionary Computation, Signal Analysis, Automatic Feature Extraction, EEG Analysis.

## 1. INTRODUCTION

In most natural and artificial environments, signals are recorded and their analysis can lead to having better knowledge about the processes involved. Thus, signal analysis and classification is a main topic in many research areas.

Traditionally, signal classification can be roughly divided into three main steps: preprocessing, feature extraction and classification. The first one, preprocessing, involves the application of different techniques in order to improve the quality of the signal. Noise filtering is an example of preprocessing. The next step is feature extraction. Its objective is to compute from the signal a set of different measures that represent the information in the best way. These features include information in one or several frequency bands, time events, etc. These features will be used as inputs in the last step, classification, which involves the use of a classification system, such as Support Vector Machines or Artificial Neural Networks.

One of the biggest problems in this process occurs in the second stage, feature extraction. Usually, the extraction of the features is done by means of manually selecting which are thought to be the best ones. Thus, previous knowledge about the signals is needed, and the features extracted are not guaranteed to be the best ones since they are based on the knowledge that the expert may or may not have. If the expert selects good features with important information, high accuracies will be obtained in the classification process. On the other side, if the features do not contain information interesting for the classification, poor accuracies will be obtained and the system will not be useful. Moreover, this feature selection process, as is manually performed by the expert, has a time and effort cost.

In this paper, a technique for signal analysis and classification is proposed. The main novelty of this technique is that no previous knowledge about the signals is necessary to extract the features. Instead of it, this technique can analyze the signals and automatically extract the features that it considers to have the most important information for the classification. Therefore, these features are not limited by the knowledge that the expert may/may not have and the classification algorithm can return higher accuracies with them as inputs.

This automatic feature extraction system also leads to having an important advantage: the new features with the knowledge for the classification are extracted without the limitation of a manual process. So, they can be analyzed in order to understand where the information for the classification is contained. These means that new knowledge about the signals the can be obtained.

The method proposed in this paper uses Genetic Programming (GP) in order to perform the automatic feature extraction and the classification in one single step.

To test the performance of this method, it was applied to an Electroencephalogram (EEG) dataset related to epilepsy disease. EEG is the recording of the electrical activity of the brain. These recordings contain valuable information for understanding the brain activity and the different brain diseases, like epilepsy.

After stroke, epilepsy is the most prevalent neurological disorder in humans. It is characterized by seizures, caused by abnormal activity of the brain. These seizures is characterized by specific detectable signal features in the EEG recordings. Therefore, the detection of these seizures in the EEG signals is very important in the diagnosis and treatment of epilepsy. This detection of seizures can be done by means of classification of the EEG signals. However, these EEG signals contain a huge amount of data and visual inspection becomes a hard manual process that involves using high qualified experts and a great amount of time. So, the developing of automatic EEG analysis techniques is of great significance for epilepsy diagnosis, treatment and understanding. Furthermore, this automatic EEG analysis technique can be useful not only for epilepsy understanding, but also for increasing the knowledge of any neurological activity or disorder.

## 2. BACKGROUND

Advances in signal processing have always had a deep impact in any knowledge area to which it was applied. A good and very well documented example is EEG processing. A wide range of techniques have been used to analyze EEGs. So, this problem is a perfect candidate in order to test the performance of new techniques and compare their results, because they are very well documented [11]. Therefore, EEG classification has become a hot topic in research. Specifically, epileptic EEG analysis has been the focus of the work of many different authors. These works have analyzed EEG signals from different points of view and domains.

Signal classification through frequency analysis is one of the most recurrent techniques. It has been successfully applied to many knowledge areas, although there are few techniques that allow its use in an automated way. The methods in frequency analysis are usually composed of three phases: selection of frequency bands, feature extraction from those bands, and classification by means of some kind of automatic method which uses the features previously extracted.

A good example is [15], where Schröeder et al. use a Genetic Algorithm (GA) to perform the selection of EEG channels. This process is followed by a feature extraction. The extracted features are used to perform the classification of the signal according to a Support Vector Machine (SVM). The paper does not propose an automatic feature extraction or frequency band selection, it only performs an automatic selection of channel in the most suitable way. In [6], the authors describe an alternative in which features are chosen trying to identify the most important ones from a predefined set (energy, fractal dimension, etc.).

It is also worth mentioning Dalponte's work because a frequency band selection is performed in an automatic way [5]. This work is based on exhaustive iterative search over every possible combination of frequency and time bands. This search is performed iteratively between the maximum and minimum values set on the algorithm. This algorithm returns only a single frequency band which is used to extract all the features for the classification step. The problem is that the information may be present on some different ranges within the returned frequency band. If specific ranges are known, the algorithm could return more accurate information and classification.

Rivero et al. in [7] perform another interesting work. In this one, the authors describe a system based on GP which was used for feature selection and classification. This method has as main weakness that it performs the classification from a single signal instead of being a more general method. It combines in a single process the feature selection and classification.

Frequency and time-based features signal are some of the most usual methods to analyze EEG. For example, Tzallas et. Al. applied a pseudo Wigner-Ville and the smoothed-pseudo Wigner-Ville distributions to extract features from the signal [18]. Those features were used as inputs to an Artificial Neural Network (ANN) that performs the classification. Another parameter could be the energy of the signal, used in [13] to perform the classification.

Although ANNs are usually the first choice to perform the classification of the signal, other works which use alternative methods can be found in literature. For example, in [12], the work uses a Welch method to perform the analysis of the frequency bands and, as classification technique, they use a decision tree based on the Welch-extracted features.

Other techniques use alternatives to frequency analysis over temporal signals to analyze and extract features. In this sense, the wavelet transform has been an important tool to perform the analysis of the signal [2]. These wavelets are able to analyze EEG signals simultaneously in time and frequency. Subasi uses the information extracted by a wavelet analysis of the EEG signals as inputs for an ANN [17]. The wavelet transform was used not only to perform a time-frequency analysis in other recent works, but also to extract information from the energy of the signal. This information was used once again as input to an ANN. In [4], the author uses a wavelet-Fourier to automatically detect the EEG seizure.

Another interesting point of view is described in [10] where the authors use a combination of different transforms (wavelets, Euclidean distance and phase-space reconstruction). The information extracted is used as inputs for a Fuzzy neural network to perform the classification of seizure signals. Alternatively to wavelets or other transforms, the authors in [9] propose the use of a PCA-Multiscale to perform the analysis and de-noising of an EEG signal. Once the signal is analyzed, they perform the classification of the seizures by means of C4.5 algorithm as classification method.

A topic that has also been used in the analysis of different signals is the concept of entropy. This is a concept based on Shannon's information theory [16]. Different estimators can be defined by using the entropy of the signal [8]. In this last work, features are extracted to be subsequently used as inputs of a neuro-fuzzy system. This system has the ability to develop an adaptive inference (ANFIS).

Finally, a different tendency is to treat EEGs like chaotic signals. For example, in [1] the authors used Lyapunov exponents and Jacobian Matrices to analyze the signal. In [19], the authors extract different features by using Lyapunov exponents. Those features were then used as inputs to classify the signal. Alternatively, other works within chaotic analysis uses fractal dimensions to analyze the signal. Then, those fractal features are used in combination with a SVM to provide the classification of the signal [14].

A remarkable point is that none of the previously mentioned works uses nothing more than single signals. This paper aims to present the behavior of the presented system by testing its performance over a signal composed of pairs of measures.

Also, those works perform classification with previously extracted features, i.e., a human expert is needed to select which of the features are the best for the classification. Few works perform feature extraction. In one of the most recent, a series of features are extracted and then an automatic process is performed to select the features from this initial set [20]. In previous works, Evolutionary Computation techniques are used to automatically extract features from signals and used for the classification. This was done either with Genetic Algorithms [21] or Genetic Programming [7]. These works are based on extracting features from the Fourier Transform. However, features from a single signal are extracted.

## 3. PROBLEM DESCRIPTION

The problem to be solved in this paper refers to the classification of epileptic EEG signals. The dataset used in this paper was described in [3]. The proposed system will be used to classify EEG signals from epileptic patients in order to discriminate epilepsy episodes from the normal ones.

The dataset contains intracranial recordings of EEGs from 5 epileptic patients. An extracranial electrode was used as reference in positions Fz and Pz over a 10/20 system. A sampling rate of 512 or 1024 Hz was used to record the EEG signals. A band-pass filter between 0.5 Hz and 150 Hz with a fourth-order Butterworth filter was applied to each EEG signal recorded. The signals recorded at 1024 Hz were down-sampled to 512 Hz.

This set of signals was divided into two subsets of signals: "focal" and "non-focal" EEG signals. Focal signals contain those channels where the first ictal EEG signal changes. Those changes were detected by a visual inspection of the two neurologists who are also board-certified electroencephalographers. Records were divided in temporal windows of 20 seconds, which involve 10240 samples for each window.

In order to build the database, 3750 pairs were randomly chosenfrom the focal EEG channels. Those 3750 pairs were recorded from signals x and y simultaneously. From these recorded signals, the records with seizure activity and those that were recorded 3 hours after the last seizure has been excluded. For each one of the pairs, one out of 5 patients was randomly selected. Once a patient was chosen, a focal EEG channel (signal x), a neighbor channel of the focal EEG for that patient (signal y) and a temporal window for the records were also chosen. Before inserting a pair of a signal, a visual inspection was performed to discard the ones which contained artifacts. In the same way, 3750 pairs of non-focal signals were chosen from non-focal EEG channels.

## 4. METHOD

In this paper, GP is used to perform the signal classification. Usually, signal classification is a process that involves the manual extraction of a series of features that are used as inputs of the classification system. However, in this work, GP is used to perform both the feature extraction and the classification at the same time.

The feature extraction is automatically done by means of an analysis in the frequency domain. Therefore, the FFT of the signals will be used instead of the originals.

In order to use GP to solve a problem, it has to be configured by specifying the nodes of the trees in the terminal and function sets. The main nodes that will be used are the following:

- MeanFFTSignal1: This node will have two children, which determine two different samples from the FFT of the first signal. This node calculates the mean of all of the FFT values between those two samples.
- StdFFTSignal1: This node will have two children, which determine two different samples from FFT of the first signal. This node calculates the standard deviation of all of the FFT values between those two samples.

These two nodes, included in the function set, operate with the FFT of the first signal. Since a pair of signals is used, two similar nodes which operate over the FFT of the second signal are needed in the set of operators. Thus, "MeanFFTSignal2" and "StdFFTSignal2" were included. The description of these nodes is similar to the previous, except that they perform the mean and standard deviation over the FFT of the second signal.

These nodes need that their children specify some numerical values that represent those samples. Therefore, new nodes have to be included in GP in order to allow it to generate these numerical values. These operators will be the basic arithmetic functions (+, -, *, %), included in the function set. The "%" operator represents protected division, which performs a normal division when the divisor is greater than 0 and returns 1 when divisor is equal to 0. Any of those operators has two children (arguments) to perform the operation.

Finally, the terminal set has to be defined, those nodes that do not have children. The terminal set includes only an ephemeral random constant between -1 and 1. This operator allows to generate random values in this range. Consequently, when it is selected to be used as a leave of a tree, a random value in that range is generated and it cannot be modified from that moment on.

In order for the terminal and function sets to meet the closure requirement and thus obtain correct trees, the evaluation of the children of the "MeanFFTSignal" or "StdFFTSignal" have to be modified. As they have been described, it is possible to have as children of these nodes subtrees that would be evaluated to real numbers, not referring to indexes of the signal. This is the more general case. Consequently, the values of the children of the "MeanFFTSignal" and the "StdFFTSignal" nodes will be transformed to be used as indexes in the following way:

1. Perform the absolute value.
2. The value is truncated so only the integer part of the number is used.
3. If the remaining positive integer value corresponds to a sample of the FFT, then return that value.
4. Else, substract to that value the length of the FFT and return to step 3.

Since GP performs a search in which a great amount of trees with random parts are generated, it is possible to obtain trees in which the frequency bands in a "MeanFFTSignal" are not constant. This can happen because a children of this node can be another "MeanFFTSignal" or "StdFFTSignal".

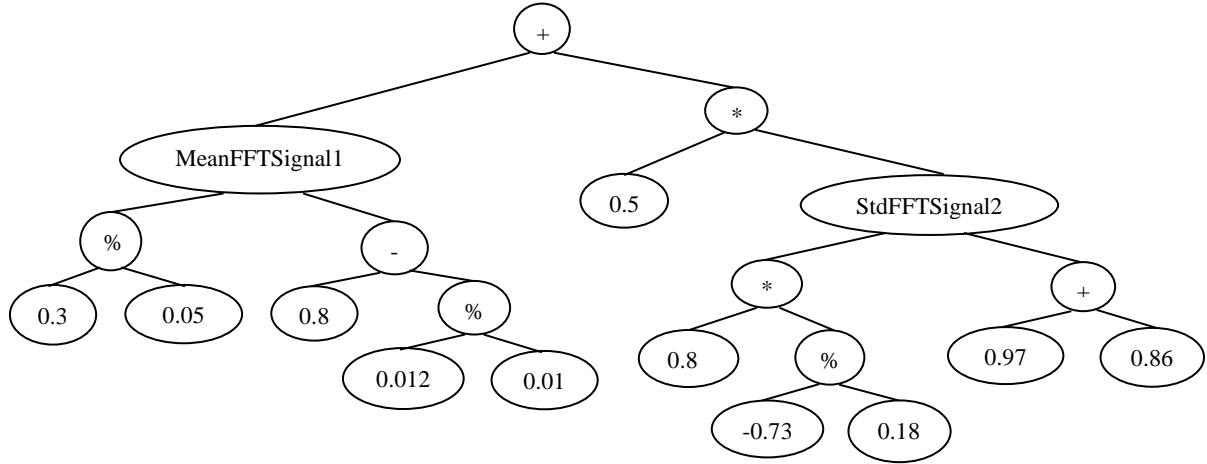

**Figure 1. Example of GP tree**

This situation is not desired, and therefore a restriction in the building of the trees is set: the "MeanFFTSignal" and the "StdFFTSignal" operators cannot be predecessor or offspring of the other MeanFFT or StdFFT operators. Therefore, the GP algorithm has to be modified to allow this restriction. The modification of the GP algorithm implies that:

- In the creation of trees or sub-trees, the "MeanFFTSignal" and "StdFFTSignal" operators cannot be chosen if one of these nodes is already a predecessor of this sub-tree. This modification was done in the creation algorithm and in the mutation algorithm.
- In the crossover of trees, when a node of one tree is selected for crossover, two situations can happen:
  - This node is in the sub-trees of the children of a "MeanFFTSignal" or "StdFFTSignal" node. In this case, the second node to be chosen for the crossover must also be part of any sub-tree of these nodes in the second tree.
  - This node is not in the sub-trees of the children of a "MeanFFTSignal" or "StdFFTSignal" node. In this case, the second node to be chosen for the crossover must not be part of any sub-tree of these nodes in the second tree.

The terminal and function sets can be seen in Table 1. All of the nodes in the function set have two children.

An example of a tree that can be built with these terminal and function sets can be seen in Figure 1. The children of the "MeanFFTSignal" node have values of 3.91 and -19.2. Therefore, the values used as sample index for this node are 3 and 19. The children of the "StdFFTSignal2 node have values of -3.41 and 1.83. Therefore, the values used as sample index for this node are 1 and 3.

Thus, that tree can be read similarly to: "take the average of the FFT of the first signal between samples 3 and 19 and add the value of the standard deviation of the FFT of the second signal between samples 1 and 3 multiplied by 0.5".

**Table 1: Terminal and function sets**

| Terminal set | Function set |
|---|---|
| [-1,1] | MeanFFTSignal1 |
|  | MeanStdSignal1 |
|  | MeanFFTSignal2 |
|  | MeanStdSignal2 |
|  | +, -, *, % |

In this example, if a FFT with 10240 samples from a 512 Hz signal was used, this tree could be read as "perform the average of the Fourier Transform of the first signal value between frequencies 0.15 Hz and 0.95 Hz, and add the standard deviation of the Fourier Transform of the second signal between frequencies 0.05 Hz and 0.15 Hz multiplied by 0.5"

The result of the evaluation of these expressions in each signal is used as input to a hyperbolic tangent function such as:

$$\tanh(x) = \frac{e^x - e^{-x}}{e^x + e^{-x}}$$

This function returns values close to 1 for positive values and close to -1 for negative values. The value of the desired class of the signal can be either 1 or -1.

The fitness of the tree is calculated as the average of the absolute value of the difference between the output of this function and the desired class for each of the signals in the training set:

$$fitness = \sum_{i=1}^{N} |t_i - o_i|$$

where $t_i$ is the target class value and $o_i$ is the obtained output for the pair of signals i.

The evolutionary process attempts to minimize the fitness function. This means that the tree should return values with a high absolute value and, with them, the tanh function will return values close to -1 or 1.

To use these expressions for classification, a threshold of 0 is used on the output to decide to which class the pair of signals belongs to. Those pair of signals with values higher than 0 will be assigned to class 1 and values lower than 0 will be assigned to class -1.

## 5. RESULTS

The technique proposed in this paper was used to classify the EEG signals described in section 3. Two different experiments were carried out: using the whole dataset for training and splitting the dataset into training, validation and test.

In the first experiment performed, the whole dataset was used as training set. The objective of this experiment was to test this system as a signal analysis technique, not trying to develop a classification system.

In this case, the training process was performed with 7500 patterns, each one from a different pair of signals. Previously, a series of experiments were run in order to set the values of the GP parameters. As result, the GP parameters were given the following values:

- Population size: 1000 individuals.
- Crossover rate: 95 %
- Mutation probability: 4%
- Selection algorithm: 2-individual tournament
- Creation algorithm: Ramped half-and-half
- Maximum tree height: 9

The GP algorithm was run until 20 generations were performed with no change in the best individual.

Once the GP parameter values were set, 50 different experiments were run. Table 2 shows the average results of the 50 independent runs carried out. The measures contained in the table are sensitivity, specificity, accuracy, area under the curve (AUC) and a 95% confidence interval. Below the accuracy value, the standard deviation of the 50 independent runs can also be seen.

**Table 2. Performance measures in training**

| Measure | Value |
|---|---|
| Sensitivity | 79.40 % |
| Specificity | 69.13 % |
| Accuracy | 74.27 %<br>1.85 |
| AUC | 87.69 % |
| CI | (86.89, 88.48) |

In the second experiment the objective was to develop a classification system. In this case, some modifications have been done to the overall process.

In a classification system the goodness cannot be measured with the accuracy in the training set, because the system is going to be used in new patterns. So, in order to properly measure the goodness, it has to be tested with new unseen patterns. The accuracy in the training set is not a valid measure of its goodness. Therefore, a different pattern set is needed to perform a test with unseen data. This pattern set is called the test set and is usually extracted from the initial data set so the training and test sets are disjoint.

A common problem in these systems is overfitting. This problem occurs when the training process returns an individual with a high accuracy, but when it is evaluated with the test set, the results are much worse. To avoid overfitting, a different pattern set is used here: validation set. This pattern set is used to control the training process.

By using this validation set, the evaluation of each individual is slightly modified: when an individual has to be evaluated, this is done with the training and validation sets independently, so a training and a validation fitness are stored with each individual. The training fitness value was used as the fitness value to guide the evolution process. However, when this process finishes, the returned individual is the one with the lowest validation fitness in the whole training process. Finally, the individual returned by the evolutionary process is used to calculate the test measures with the test set.

Therefore, the pattern set was randomly split into three different non-overlapping sets: training, validation and test. Each of these sets had 33% of the data set, i.e., 2500 patterns.

This schema was used 50 times in 50 different independent executions. For running these experiments, the same GP parameter configuration was used. As done before, the GP algorithm was run until no change was done in the best individual in 20 successive generations.

As result, Table 3 shows the same values as Table 2 (sensibility, specificity, accuracy, AUC and confidence interval), but, this time, the values shown are the average measures for the 50 runs of the test datasets. Below the accuracy value in the test set, the standard deviation of the 50 independent executions can also be seen.

**Table 3. Performance measures in test**

| Measure | Value |
|---|---|
| Sensitivity | 79.60 % |
| Specificity | 63.23 % |
| Accuracy | 70.69 %<br>1.63 |
| AUC | 85.32 % |
| CI | (83.77, 86.86) |

As can be expected, the results obtained with the test sets are slightly lower to those obtained with the whole dataset used for training. The mean accuracy in test was 70.69%. Even this accuracy may seem quite low for a medical application, this technique performs a very simple analysis in the frequency domain, based on the Fourier Transform. With a more complicated analysis, such as wavelets, these results could be improved.

To the knowledge of the authors, no other works have been published that reports accuracy values with the same epilepsy database.

## 6. CONCLUSIONS

This paper describes a technique for signal classification in which a pair of signals is used by this system to perform the classification. As opposed to most of the signal classification systems, this technique can automatically extract features from both signals in order to perform the classification. This is the main advantage of the proposed system: it is not necessary to perform a previous feature extraction process. As shown in section 4, this system does not need an expert to specify the features to be used as inputs for the classification. Instead of it, the system extracts the features that it considers are the most representative to solve the problem. As a consequence, the automatically extracted features are expected to perform better than those manually extracted because they are not based on previous (and possibly) incomplete knowledge.

Regarding the practical application, this technique is promising. As section 5 shows, the system proposed here is able to analyze and classify EEG signals in a very complex problem. Higher accuracies can be obtained if a more complex analysis is performed. With this in mind, a system for automatic epileptic seizure detection with no participation of the human expert could be developed. This system would not need an expert to perform visual inspection of the signals, or to have previous knowledge in order to decide the features to be used. This system could also be used not only in epileptic seizure detection but also in any other EEG processing problem.

In this case, this system has been applied to an epileptic seizure detection problem, but it could be applied to any other signal classification problem.

## 7. FUTURE WORKS

In this work, GP has been used to analyze signals in the frequency domain. This is the most basic type of signal analysis, used in many studies. The results reported in this paper could be improved with other types of analysis, such as wavelets or entropies. The use of this technique mixing GP and other analysis can lead to having higher accuracies.

## 8. ACKNOWLEDGMENTS

The development of the experiments described in this work was performed on computers from the Supercomputing Center of Galicia (CESGA).

This work is supported by the General Directorate of Culture, Education and University Management of Xunta de Galicia (Ref. GRC2014/049, R2014/039 and CN2012/211) and the European Fund for Reginal Development (FEDER) in the European Union "Collaborative Project on Medical Informatics (CIMED)" Pl13/00280 funded by the Carlos III Health Institute.

## 9. REFERENCES

[1] Abarbanel, H.D., Brown, R., and Kennel, M., 1991. Lyapunov exponents in chaotic systems: their importance and their evaluation using observed data. *International Journal of Modern Physics B 5*, 09, 1347-1375. DOI= http://dx.doi.org/10.1142/S021797929100064X.

[2] Addison, P.S., 2002. *The illustrated wavelet transform handbook: introductory theory and applications in science, engineering, medicine and finance*. CRC Press.

[3] Andrzejak, R.G., Schindler, K., and Rummel, C., 2012. Nonrandomness, nonlinear dependence, and nonstationarity of electroencephalographic recordings from epilepsy patients. *Physical Review E 86*, 4, 046206. DOI= http://dx.doi.org/10.1103/PhysRevE.86.046206.

[4] Chen, G., 2014. Automatic EEG seizure detection using dual-tree complex wavelet-Fourier features. *Expert Systems with Applications 41*, 5, 2391-2394.

[5] Dalponte, M., Bovolo, F., and Bruzzone, L., 2007. Automatic selection of frequency and time intervals for classification of EEG signals. *Electronics Letters 43*, 25, 1406-1408. DOI= http://dx.doi.org/10.1049/el:20072428.

[6] Deriche, M. and Al-Ani, A., 2001. A new algorithm for EEG feature selection using mutual information. In *Acoustics, Speech, and Signal Processing, 2001. Proceedings.(ICASSP'01). 2001 IEEE International Conference on* IEEE, 1057-1060. DOI= http://dx.doi.org/10.1109/ICASSP.2001.941101.

[7] Fernández-Blanco, E., Rivero, D., Gestal, M., and Dorado, J., 2013. Classification of signals by means of Genetic Programming. *Soft Computing 17*, 10, 1929-1937. DOI= http://dx.doi.org/10.1007/s00500-013-1036-4.

[8] Kannathal, N., Choo, M.L., Acharya, U.R., and Sadasivan, P., 2005. Entropies for detection of epilepsy in EEG. *Computer methods and programs in biomedicine 80*, 3, 187-194. DOI= http://dx.doi.org/10.1016/j.cmpb.2005.06.012.

[9] Kevric, J. and Subasi, A., 2014. The Effect of Multiscale PCA De-noising in Epileptic Seizure Detection. *Journal of medical systems 38*, 10, 1-13.

[10] Lee, S.-H., Lim, J.S., Kim, J.-K., Yang, J., and Lee, Y., 2014. Classification of normal and epileptic seizure EEG signals using wavelet transform, phase-space reconstruction, and Euclidean distance. *Computer methods and programs in biomedicine 116*, 1, 10-25.

[11] Mohseni, H.R., Maghsoudi, A., and Shamsollahi, M.B., 2006. Seizure detection in EEG signals: A comparison of different approaches. In *Conf Proc IEEE Eng Med Biol Soc* Citeseer, 6724-6727. DOI= http://dx.doi.org/0.1109/IEMBS.2006.260931.

[12] Polat, K. and Güneş, S., 2007. Classification of epileptiform EEG using a hybrid system based on decision tree classifier and fast Fourier transform. *Applied Mathematics and Computation 187*, 2, 1017-1026. DOI= http://dx.doi.org/10.1016/j.amc.2006.09.022.

[13] Rivero, D., Dorado, J., Rabuñal, J., and Pazos, A., 2009. Evolving simple feed-forward and recurrent ANNs for signal classification: A comparison. In *Neural Networks, 2009. IJCNN 2009. International Joint Conference on* IEEE, 2685-2692. DOI= http://dx.doi.org/10.1109/IJCNN.2009.5178621.

[14] Schneider, M., Mustaro, P.N., and Lima, C.a.M., 2009. Automatic recognition of epileptic seizure in EEG via support vector machine and dimension fractal. In *Neural Networks, 2009. IJCNN 2009. International Joint*

Conference on IEEE, 2841-2845. DOI= http://dx.doi.org/10.1109/IJCNN.2009.5179059

[15] Schroder, M., Bogdan, M., Hinterberger, T., and Birbaumer, N., 2003. Automated EEG feature selection for brain computer interfaces. In *Neural Engineering, 2003. Conference Proceedings. First International IEEE EMBS Conference on* IEEE, 626-629. DOI= http://dx.doi.org/10.1109/CNE.2003.1196906.

[16] Shannon, C.E., 2001. A mathematical theory of communication. *ACM SIGMOBILE Mobile Computing and Communications Review 5*, 1, 3-55. DOI= http://dx.doi.org/10.1002/j.1538-7305.1948.tb00917.x.

[17] Subasi, A., 2007. EEG signal classification using wavelet feature extraction and a mixture of expert model. *Expert Systems with Applications 32*, 4, 1084-1093. DOI= http://dx.doi.org/10.1016/j.eswa.2006.02.005.

[18] Tzallas, A.T., Tsipouras, M.G., and Fotiadis, D.I., 2009. Epileptic seizure detection in EEGs using time–frequency analysis. *Information Technology in Biomedicine, IEEE Transactions on 13*, 5, 703-710. DOI= http://dx.doi.org/10.1109/TITB.2009.2017939.

[19] Übeyli, E.D., 2010. Lyapunov exponents/probabilistic neural networks for analysis of EEG signals. *Expert Systems with Applications 37*, 2, 985-992. DOI= http://dx.doi.org/10.1016/j.eswa.2009.05.078.

[20] Jenke, R., Peer, A., Buss, M., 2014. Feature Extraction and Selection for Emotion Recognition from EEG. *Affective Computing, IEEE Transactions on*, 5, 3, 327,339, July-Sept. 1 2014

[21] Rivero, D., Aguiar-Pulido, V., Fernandez-Blanco, E., Gestal, Marcos, 2013. Using genetic algorithms for automatic recurrent ANN development: an application to EEG signal classification. International Journal of Data Mining, Modelling and Management, 5, 182-191. DOI= http://dx.doi.org/10.1504/IJDMMM.2013.053695